\documentclass[11pt]{article}
\usepackage{amsmath, amsfonts, amssymb}
\usepackage{amsthm}
\usepackage{authblk, color, bm, graphicx}
\usepackage[left=1in,top=1.2in,right=1in,bottom=1.2in]{geometry}
\usepackage{xspace}
\usepackage{hyperref}

\usepackage{balance}
\usepackage{subcaption}
\usepackage{multirow}
\usepackage[linesnumbered,ruled,vlined]{algorithm2e}
\SetKwInput{KwInput}{Input}                
\SetKwInput{KwOutput}{Output}              

\title{Risk-Aware Bid Optimization for Online Display Advertisement \footnote{This article was reviewed and accepted in Proceedings of the 31st ACM International Conference on Information and Knowledge Management (CIKM'22) \href{https://doi.org/10.1145/3511808.3557436}{DOI}
.}}

\author[1]{Rui Fan\footnote{Corresponding author, 
email addresses: 
\textit{rui.fan@hec.ca}
}}
\author[1]{Erick Delage \footnote{Email addresses: \textit{erick.delage@hec.ca}}}

\affil[1]{GERAD \& Department of Decision Sciences, HEC Montr\'eal, Montr\'eal, QC H3T 2A7, Canada.}

\date{}

\begin{document}

\newtheorem{ex}{Example}
\newtheorem{thm}{Theorem}
\newtheorem{assumption}{Assumption}
\newtheorem{remark}{Remark}
\newtheorem{lemma}{Theorem}

\def\Expect{{\mathbb E}}
\def\Prob{{\mathbb P}}
\def\minimize{\mathop{\rm minimize}}
\def\maximize{\mathop{\rm maximize}}
\def\subto{{\rm subject \mbox{   }\rm to}}
\def\min{\mathop{\rm min}}
\def\max{\mathop{\rm max}}
\def\argmax{\mathop{\rm argmax}}
\def\argmin{\mathop{\rm argmin}}

\newcommand{\alphap}{\alpha^\prime}
\newcommand{\bRNP}{{b^{\mbox{rnp}}}}
\newcommand{\tbRNP}{{\tilde{b}^{\mbox{rnp}}}}
\newcommand{\bRAP}{{b^{\mbox{rap}}}}
\newcommand{\tbRAP}{{\tilde{b}^{\mbox{rap}}}}
\newcommand{\Gain}{\mathcal{G}}
\newcommand{\barB}{\bar{B}}
\newcommand{\quoteIt}[1]{``#1''}
\newcommand{\removed}[1]{}

\maketitle

\abstract{This research focuses on the bid optimization problem in the real-time bidding setting for online display advertisements, where an advertiser, or the advertiser's agent, has access to the features of the website visitor and the type of ad slots, to decide the optimal bid prices given a predetermined total advertisement budget. We propose a risk-aware data-driven bid optimization model that maximizes the expected profit for the advertiser by exploiting historical data to design upfront a bidding policy, mapping the type of advertisement opportunity to a bid price, and accounting for the risk of violating the budget constraint during a given period of time. After employing a Lagrangian relaxation, we derive a parametrized closed-form expression for the optimal bidding strategy. Using a real-world dataset, we demonstrate that our risk-averse method can effectively control the risk of overspending the budget while achieving a competitive level of profit compared with the risk-neutral model and a state-of-the-art data-driven risk-aware bidding approach.
\\
}


\section{Introduction}
In a real-time-bidding (RTB) online display advertisement setting, the advertiser is given the website traffic of clients and ad slots with many different features, including logs, regions, ad slots format, etc., in order to decide a bidding policy for ads potentially displayed to a wide range of future users. The bidding process in digital advertising \cite{Auction} is based on the second-price auction \cite{secondauction} model, where advertisers or their agents win the auction of a given opportunity if they offer the highest bid price among the competitors and pays the second highest bidder's bid price. The bidders in the auction can be the advertiser themselves or their agents, which we will refer to as the decision makers for the bidding policy. 

These bidding policies typically need to be made with incomplete information about the potential click-through rate (CTR)\footnote{The click-through rate of an advertisement is the number of times a click is made on the advertisement, divided by the number of times the ad is shown, which is also called the number of impressions \cite{wiki}.}, winning prices\footnote{The winning price (market price) is the price that is paid for the ad spot.}, the net value of each customer to the company, etc. Instead, they rely on historical data that could be shared by the advertisement platform or accumulated internally by advertisers themselves. The decision makers in this problem need to use this historical data to estimate these critical values and develop the optimal bidding policy for the forthcoming bidding opportunities.

Many companies around the world exploit online marketing, and in 2020, the total spending on digital ads worldwide reached 378.16 billion US\$ \cite{market}. Yet, effective use of online marketing budgets requires robust statistical methods that can identify key interactions among a possibly biased historical dataset and algorithms that efficiently converge to optimal risk-aware bidding policies. Moreover, the currently available methods seldom consider the risk in the bidding activity and, to the best of our knowledge, have not explicitly addressed the risk of incurring excessive expenses. 

In reality, the marketing budget is determined ahead and depends on a certain period of time. The budget is given at the beginning of the period when the decision makers have no knowledge about what opportunities will precisely realize. If they do not account for the risk of going over budget, valuable opportunities that appear at a later time will likely be missed. Therefore, being able to identify bidding policies that control the risk of excessive expense during a given period of time is of practical concern.

This paper considers an advertiser who needs to design a bidding policy, mapping the type of advertisement opportunity to a bid price, that will be deployed over a period of time while effectively controlling the risk of violating a specified budget constraint. For this purpose, we propose a static risk-aware bid optimization model that can exploit historical data to prescribe an optimal bidding policy. The approach is hybrid in how it considers a stochastic model that mixes both an empirical distribution to model the type of opportunities, and a parametric distribution to model the winning price and click potential when formulating the problem. This allows us to derive novel closed-form expressions for the optimal risk-neutral and risk-averse bidding policies. The optimal bidding policies are easy to implement and interpret given that they involve simple functions of the estimated conditional CTR and conditional mean and variance of the anticipated winning price. To control the risk of running out of budget, we employ an expected utility model that can account for the advertiser's risk aversion.
Finally, the proposed bidding strategies are implemented and evaluated on a real-world dataset. These experiments \footnote{The code of experiments is available at \url{https://github.com/ReneeRuiFAN/risk-aware_bid_optimization}} 
show that our bidding policy effectively controls the risk of expense going beyond the budget while outperforming state-of-the-art risk-aware bidding strategies in terms of the average profit that is achieved.

The rest of the paper is divided as follows. Section \ref{sec:litrev} presents our review of the literature. Section \ref{sec:SM} proposes a stochastic model for describing a random bid opportunity offered to an advertiser. Sections \ref{sec:rnp} and \ref{sec:rap} respectively present our so-called risk-neutral and risk-averse bid optimization problems. Finally, sections \ref{sec:exp} and \ref{sec:num} respectively describe our experiment design and numerical results, while Section \ref{sec:conc} concludes.

\section{Literature Review}\label{sec:litrev}

Our literature review covers the three main topics of this paper. 

\subsection{CTR and winning price prediction}

The CTR prediction is a binary classification problem, commonly used estimators in computational advertising are linear-based \cite{LR}, tree-based \cite{GBRT}, and many models have been developed based on the Factorization Machines (FM) \cite{FM, FMctr, SFM, FFM, FwFM} which can better fit the feature combination and sparse data often found in the display advertising datasets. In recent years, with the development of Deep Learning (DL) research and recommendation systems, many researchers have applied Deep Neural Network (DNN) based models \cite{CCPM, PIN, widedeep}. The idea of combining the DNN and FM is widely accepted in the research and represents the leading performance in real-world usage \cite{guo2017deepfm, FiBiNET, FEFM}. 

The winning price prediction is also called the bid landscape problem. \cite{bidlands} uses the gradient boosting decision trees to model the winning price. \cite{w_censor} proposed the censored regression model to deal with the problem when some historical winning prices are unknown to the advertiser. The recent DL advancement also applies to the bid landscape problem. \cite{deepcensor} proposed the Deep Censored Learning model that uses a DL model for CTR prediction to boost the prediction quality on the winning price and considers its distribution into the learning. Deep Landscape Forecasting (DLF) model \cite{ren2019deep} combines DL for probability distribution forecasting and survival analysis for censorship handling based on a recurrent neural network (RNN) to model the conditional winning probability with respect to each bid price. 

\subsection{Real-Time Bidding Strategies}

Linear-based strategies are commonly used in developing bidding strategies. \cite{Auction,rtb_performance} proposed bidding policies linearly-related with the estimation of the value of click, so-called truthful bidding, while \cite{BO_inventory} constructed bid prices that depend on the predicted CTR. Besides the profit maximization objective, a dual-based bidding framework \cite{liu2017dual} derived from a strict second-price auction assumption is generally applicable to the multiple ads scenario with various objectives and constraints. \cite{Yang_2019} studied the common case where advertisers aim to maximize the number of conversions, and set cost-per-click (CPC) as a constraint. 

In \cite{orbt}, the authors introduced a non-linear bidding strategy model (called ORTB) with the estimated CTR as the input of the bidding function that tries to bid on more impressions rather than focus on a small set of high-value impressions. The paper \cite{J_ortb} solves the bid problem in cases where impressions are generated by homogeneous Poisson processes and winning prices are independent and identically distributed (i.i.d). \cite{bidmachine} model CTR learning and winning price estimation as part of bid optimization for campaign profit maximization as a whole and perform a joint optimization.

Researchers also consider bid optimization in a multi-stage setting as a sequential decision process, where Reinforcement Learning (RL) can play an important role. Indeed, the bid optimization problem can be formulated by Markov Decision Process (MDP), where the bid prices are the actions and the realized clicks provide rewards to the RL agent. 
In this RL model, the leftover budget can be integrated into the state space, such as in \cite{Markov, Cai_RL}, or used to influence the reward in \cite{RL_budget}.
Researchers have also formulated the problem using a multi-agent RL framework \cite{RL_agent, RL_search}. However, all these RL approaches are generally more computationally expensive to solve compared with static models. Moreover, their solutions usually lack interpretability.

\subsection{Risk-Aware Bidding Strategies}

Based on a multi-stage problem setting, many researchers looked into the feedback control problem during the bidding process \cite{Frontierbid}, which controls the risk of unstable performance and keeps the optimization process along the stages using a dynamic system. \cite{bidrnn} add a penalty to the cost, if the bidding policy falls short of its key performance indicators to improve the robustness of performance under uncertainty. \cite{randombid} introduced a bid randomization mechanism to help exploration in a partially observed market and control the uncertainty in the auction-based bidding process.

Most closely related to our work, \cite{Zhang_2017} proposed the risk management on profit (RMP) model that also models the bidding process as a static problem. The model focuses on controlling the risk of the generated profit, while, in sharp contrast with our model, it does not address the risk of expense going over the budget. The authors also assume that profit risk is solely caused by CTR estimation error, which they model using Bayesian logistic regression, and end up over-simplifying the problem by assuming that the winning price is independent of the type of advertisement opportunity. 


In the display advertisement field, the utility of bidding is often defined as the profit of clicks. The papers \cite{utilitybidder, utilityauction, Zhang_2017} define the value of click $v$ as the value of the sum of winning prices divided by total clicks and measure utility as the profit of bidding which is the difference between the total value of clicks and the expense paid. 
In these papers, the authors use expected utility theory \cite{expect_propose, expect_theorem}. 
The exponential utility function, which accounts for a constant absolute risk aversion, is probably the most commonly used utility function. It can also be interpreted as employing an entropic risk measure \cite{entropicrisk}, which satisfies the axioms of convex risk measures \cite{coherent}. 

\section{Stochastic Model and Training}\label{sec:SM}

For each bidding opportunity with an observable feature vector $X$ that represents both user and ad information, the bidding optimization problem will account for three dependent random variables: the realized click $C$ represents if the fact is that the ad gets clicked, the winning price $W$, and the net value of the customer to the company $V$. Also, we make the following assumption to facilitate the modeling of $C, W, V$ given $X$. These dependency assumptions are commonly used in other literature \cite{orbt,Zhang_2017,bidlands,bidmachine}. 
\begin{assumption}\label{ass:indep}
The winning price $W$, realized $C$, and the net value of the customer $V$ are mutually independent given $X$.
\end{assumption}
Additionally, our approach will optimize a bidding policy over a batch of $M$ opportunities, which are assumed i.i.d. This will simulate a one-shot decision situation where a bidding policy needs to be defined in order to run for a given period of time under a fixed budget. 
Table \ref{tab:variable} summarizes the definitions of the variables used in our formulation, which are discussed next.

\begin{remark}
The idea of considering a batch of $M$ opportunities is a distinctive feature of our decision models. Previous static approaches (see \cite{Zhang_2017, orbt}) usually assume that $M$ is large enough for the law of large numbers to apply and impose that the expected expense be smaller than the average budget. In \cite{Zhang_2017}, for example, the proposed bidding policy is simply scaled to the same extent that gives the maximum overall profit under the predetermined budget during the hyperparameters tuning. Our models instead account for the fact that $M$ can in practice be too small for the average expense to have converged to its expected value. We will however exploit a property of the entropic risk measure that enables us to reduce a batch problem of size $M>1$ to an equivalent instantaneous $M=1$ problem, which greatly simplifies the analysis. 
\end{remark}

In the rest of this section, we propose conditional models for CTR, winning price, winning probability, and value of the customer.

\subsection{Modeling Conditional CTR}

We assume that the CTR depends on the opportunity's features $X$, and formally denote: $\theta(X):=\mathbb{P}(C|X)$.
In the context of this work, we will employ a DeepFM model \cite{guo2017deepfm} to estimate $\theta(X)$. We note that this choice is not limiting and that other CTR prediction models could also be employed if one can improve the accuracy.

\subsection{Modeling Conditional Winning Price Distribution}

We assume that conditional on observing $X$, the winning price $W$ follows the normal distribution
\begin{math}
  W \sim N (\hat{w} (X), \sigma(X)) ,
\end{math}
where the conditional standard deviation $\sigma(X)$ also depends on the given opportunity's features $X$, so that we have the parametrized probability distribution function of the winning price $W$ modeled as follows:
\begin{displaymath}
  f_{W|X}(w) = \frac{1}{\sigma(X) \sqrt{2 \pi}} e^{-\frac{1}{2}\left(\frac{w-\hat{w}(X)}{\sigma(X)}\right)^2} .
\end{displaymath}

Therefore, to model the distribution of the winning price, we need estimators for the conditional mean $\hat{w}(X)$ and the conditional standard deviation $\sigma(X)$.

Given a dataset containing observed $(X,W)$ pairs, we can obtain an estimator of the expected winning price, conditioned on $X$, by running the regression model:
\begin{displaymath}
  \hat{w}(X):=\arg\min_{w\in\mathcal{W}} \Expect[(W-w(X))^2]\,,
\end{displaymath}
where $\mathcal{W}$ is the set of estimation functions modeled by a certain DNN architecture, and where $\Expect$ refers to the expected value under the empirical distribution observed in the dataset.

The same dataset can also be used to train an estimator of $\sigma(X)$ following a method introduced in \cite{sigma_residual}. Conceptually, we consider the residual $Z$ defined as: $Z:= (W-\hat{w}(X))^2$,
which depends on the estimator $\hat{w}(X)$, as well as the observed $W$. We can approximate the conditional variance of the winning price $\sigma^2(X)$ as the expected value of the residual $Z$. 
\begin{displaymath}
    \hat{z}(X):=\arg\min_{z\in\mathcal{Z}} \Expect[(Z-z(X))^2] ,
\end{displaymath}
where again $\mathcal{Z}$ is the set of estimation functions modeled by a certain DNN architecture. Finally, an estimator of the conditional standard deviation of winning price $\sigma(X)$\label{sec:sigma} can be obtained by: $\sigma(X) := \sqrt{\max(\hat{z}(X),\epsilon)}$,
for some small $\epsilon>0$, which ensures that the variance estimate is always positive.

In our experiments, similarly to the CTR estimator $\theta(X)$, the estimators of mean of winning price $\hat{w}(X)$ and standard deviation of winning price $\sigma(X)$ are DeepFM models. 

\subsection{Modeling Conditional Winning Probability}

Following the second price auction \cite{Auction} process, we assume that the advertisers or their agents can win the bid if they offer a bid price that is larger than the winning price. Since the expense only happens when the advertiser wins the bid, to model uncertainty about the expense, we need to model the probability of winning the bid which depends on the bid price and the winning price. For this purpose, we define a function that indicates whether the bid price wins the auction:
\begin{displaymath}
s(b, W) := 1 \{b \geq W \} =
\begin{cases}
1& b \geq W\\
0& \text{otherwise}
\end{cases}
\end{displaymath}
where $b$ is the bid price and $W$ is the winning price. Hence the conditional winning probability given $X$ can be obtained from $\Expect[s(b,W)|X]$, which depends on both the bid price $b$ and the conditional winning price distribution given $X$.

\begin{remark}
It is important to note that both the ORTB approach \cite{orbt} and the RMP approach \cite{Zhang_2017} assume that the winning price is independent of the opportunity's features $X$. The ORTB approach obtains the winning probability using a certain parametrized reciprocal function and tunes its parameter to best fit the winning probability curve as a function of the bid price. In the case of the RMP approach, the winning probability is simplistically estimated by assuming that the winning price is independent and using its empirical distribution.  
In contrast, we assume that the winning price's distribution depends on the given opportunity $X$, which is a more natural and logical assumption as advertisers are willing to bid more for the more valuable opportunities.
\end{remark}

\subsection{Modeling Conditional Value of Customer}

Ideally, if the decision maker identifies a customer who could bring a higher value to the company from the observable features vector $X$, they will be willing to bid at a higher price to improve the probability of winning the auction. Therefore, we model the value of a customer $V$ conditionally on the type of opportunity $X$: $\hat{V}(X):=\Expect[V|X]$. 
However, we note that, in our experiments, $\hat{V}(X)$ is considered as a known constant (the same value from \cite{Zhang_2017, orbt}) for simplicity. 

\begin{table}
    \caption{Variable definitions}
    \centering
    \label{tab:variable}
    \begin{tabular}{cl}
    \hline
         Variables & Description\\
    \hline
         $V$ & net value of customer for opportunity \\ 
         $\hat{V}(X)$ & conditional net value of customer estimator given $X$ \\ 
         $X$ &  observed features of opportunity \\ 
         $C$ &  indicates of click for opportunity \\ 
         $\theta(X)$ &   conditional CTR estimator given $X$\\ 
         $b(X)$ &  bid price for opportunity given $X$  \\
         $W$ &  winning price for opportunity  \\ 
         $\hat{w}(X)$ & \vtop{\hbox{\strut conditional expected winning price for }\hbox{\strut opportunity given $X$}} \\
         $s(b,W)$ &  indicates of winning the auction given $b$ and $W$ \\ 
         $\sigma(X)$ & \vtop{\hbox{\strut conditional standard deviation of winning}\hbox{\strut price for opportunity given $X$}} \\
         $M$ &  number of opportunities in a batch\\
    \hline
    \end{tabular}
\end{table}

\section{Risk-neutral Problem}\label{sec:rnp}

In this section, we first introduce a static bid optimization model that captures a risk-neutral attitude regarding the possibility of exceeding the predefined budget. While our original model considers the performance of the bidding policy over a batch, we show how this problem can be equivalently reduced to its \quoteIt{instantaneous} version (i.e. $M=1$). We then establish that by employing a classical Lagrangian relaxation, our optimal bidding policy has a closed-form expression involving the conditional statistics of CTR, winning price, and customer value.

\subsection{Risk-neutral Problem Formulation}

Considering a random batch of $M$ i.i.d. opportunities denoted by $\{(X_i, W_i, C_i, V_i)\}_{i=1}^M$, with each $V_i$, $C_i$, and $W_i$ mutually independent given $X_i$ (as per Assumption \ref{ass:indep}), the Risk-neutral Problem (RNP) seeks a bidding policy that maximizes the expected profit generated over this batch while ensuring that the total budget will be satisfied in expectation. This problem takes the following form:
\begin{eqnarray}
\bRNP(\cdot):= \argmax_{b:\mathcal{X}\rightarrow \Re^+} && \Expect[\mbox{Batch profit}] \nonumber \\
\subto &&\Expect[\mbox{Batch expense}] \leq BM  \nonumber, 
\end{eqnarray}
where $b(\cdot)$ is a bid price policy that will be employed over a batch of $M$, while $B$ captures the average budget per opportunity. Furthermore, we have that:
\begin{align*}
&\text{Batch profit} = \sum_{i=1}^M V_i C_i s(b(X_i), W_i) - \sum_{i=1}^M W_i s(b(X_i), W_i),\\
&\text{Batch expense} = \sum_{i=1}^M W_i s(b(X_i), W_i).
\end{align*}

Based on the linearity of expectation, batch expressions can be simplified. For example, we can simplify the batch profit to expected instantaneous profit per opportunity format by:
\begin{align*}
& \Expect\left[\sum_{i=1}^M V_i C_i s(b(X_i),W_i)\right]- \Expect\left[\sum_{i=1}^M W_i s(b(X_i),W_i)\right] \\
&= M \left( \Expect[VCs(b(X),W)] - \Expect[Ws(b(X),W)] \right) ,
\end{align*}
and similarly for the batch expense formula.

In the rest of this subsection, we exploit Assumption \ref{ass:indep} and the normality assumption for $W$ to obtain closed-form expressions for both the objective and constraint.

\subsubsection{Reducing expected instantaneous revenue expression}

We start by rewriting the expected instantaneous revenue expression as:
\begin{align*}
\Expect[VCs(b(X),W)]
&=\Expect[\hat{V}(X)\theta(X) \Expect[s(b(X),W)|X]] ,
\end{align*}
where we exploited (see Assumption \ref{ass:indep}) the fact that $V$, $C$, and $W$ are mutually independent of each other given $X$.

The conditional winning probability 
can be further reduced since we assume $W$ is normally distributed conditionally on $X$:
\begin{displaymath}
\Expect[s(b,W)|X]=\int_{W\leq b} f_{W|X}(W) dW 
=\Phi\left(\frac{b-\hat{w}(X)}{\sigma (X)}\right) ,
\end{displaymath}
where $f_{W|X}(w)$ is the probability density function of the winning price $W$ given $X$, and $\Phi(\cdot)$ is the cumulative distribution function of a standard normal distribution. 

Therefore, we can calculate the expected revenue given $X$ using: 
\begin{eqnarray} \label{eq:rn:revenue}
\Expect[VCs(b(X),W)|X]=\hat{V}(X) \theta(X) \Phi\left(\frac{b-\hat{w}(X)}{\sigma (X)}\right).
\end{eqnarray}

\subsubsection{Reducing expected instantaneous expense expression}

In the case of the budget constraint, we have that:
\begin{align} \label{eq:reduced:expense}
\Expect&[Ws(b,W)|X] = \int_{W\leq b} W  f_{W|X}(W)dW \nonumber \\
&=\int_{w_s\leq \frac{b-\hat{w}}{\sigma(X)}} (\sigma(X) w_s 
+\hat{w}(X)) \phi(w_s)d w_s \nonumber \\
&=\sigma(X) \int_{w_s\leq \frac{b-\hat{w}(X)}{\sigma(X)}} w_s \phi(w_s)d w_s +\hat{w}(X) \int_{w_s\leq \frac{b-\hat{w}}{\sigma(X)}} \phi(w_s)d w_s \nonumber \\
&= g(b,X):=\hat{w}(X)\Phi\left(\frac{b-\hat{w}(X)}{\sigma(X)}\right)-\sigma(X)\phi\left(\frac{b-\hat{w}(X)}{\sigma(X)}\right) ,
\end{align}
where $w_s$ follows the standard normal distribution, and $\phi(\cdot)$, $\Phi(\cdot)$ are respectively the density function and cumulative distribution function of a standard normal distribution.

\subsection{Optimal Solution for RNP Model}
After we derive the reduced revenue \eqref{eq:rn:revenue} and expense \eqref{eq:reduced:expense} expressions, we can reformulate the model for $\bRNP$ using the reduced forms:

\begin{align*}
\bRNP(\cdot) := \argmax_{b:\mathcal{X}\rightarrow \Re^+} \;\;& \Expect\left[\hat{V}(X) \theta(X) \Phi\left(\frac{b(X)-\hat{w}(X)}{\sigma (X)}\right) - g(b(X),X)\right] \nonumber \\
\subto \;\;& \Expect[g(b(X),X)] \leq B .\nonumber
\end{align*}

In an attempt to solve this problem, one can introduce the Lagrangian coefficient $\lambda\geq 0$ to obtain a relaxation of this risk-neutral profit maximizing problem:
\begin{align*}
\tbRNP_\lambda(\cdot) := \argmax_{b:\mathcal{X}\rightarrow \Re^+} \quad & \Expect\left[\hat{V}(X)\theta(X)\Phi \left(\frac{b(X)-\hat{w}(X)}{\sigma (X)}\right) - g(b(X),X)\right] \\
& - \lambda\Expect[g(b(X),X) - B]\\
=\argmax_{b:\mathcal{X}\rightarrow \Re^+} \quad &\Expect[\Gain_{\lambda}(b(X),X)] ,
\end{align*}
where
\[\Gain_{\lambda}(b,X) :=   \hat{V}(X)\theta(X)\Phi \left(\frac{b-\hat{w}(X)}{\sigma (X)}\right) - g(b,X) - \lambda (g(b,X) - B) .\]

The optimal bid price $\bRNP$ can be approximated using $\tbRNP_{\lambda^*}$ with $\lambda^*$ as the smallest value of $\lambda\geq 0$ such that $\Expect[g(\tbRNP_{\lambda^*}(X),X)] \leq B$.
Furthermore, $\lambda^*$ can be found using the bisection method.

We next provide a closed-form solution for $\tbRNP_\lambda$ in the form of Lemma \ref{thm:closedform:rnp} (see complete proof in Appendix \ref{app:rnp:thm:proof}).

\begin{lemma}\label{thm:closedform:rnp}
For any $\lambda\geq 0$, a maximizer of the Lagrangian relaxation takes the form:
\[\tbRNP_{\lambda}(X):=\arg \max_{b\in\left\{0, \; \frac{\hat{V}(X)\theta(X)}{\lambda+1},\; \infty\right\}} \Gain_\lambda(b, X),\;\forall X\in\mathcal{X} .\]
\end{lemma}

Therefore, we can that the optimal bid price $\tbRNP_{\lambda^*}$ is proportional to the estimated value of the customer and CTR given  $\lambda^*$. 

\section{Risk-averse Problem}\label{sec:rap}
The distinguishing point of the Risk-averse Problem (RAP) is that we take the risk of going over budget into consideration. We develop the risk-averse bid optimization model $\bRAP$, which maximizes the return of bidding while controlling the risk of violating the total budget for the batch of opportunities.

\subsection{The Risk-averse Budget Constraint}\label{sec:rabc}
We introduce the exponential utility function to model risk aversion in the budget constraint. Namely, we replace the expected expense constraint with:
\[\Expect[u_\alpha((1/M)\mbox{Batch expense})] \geq u_\alpha(B) ,\]
where $u_\alpha(y):= -\exp(\alpha y)$ is a concave utility function that allows the decision maker to control risk exposure using the parameter $\alpha$. 
Based on \cite{entropicrisk}, this risk-averse budget constraint can be interpreted as imposing an upper bound of $B$ on the entropic risk of the average expense in the batch:
\begin{eqnarray} \label{eq:rho}
\rho((1/M)\mbox{Batch expense}) \leq B\,,
\end{eqnarray}
where the entropic risk measure $\rho$ is a well-known convex risk measure. Furthermore, this risk-averse constraint is known to reduce to the risk-neutral one when $\alpha\rightarrow 0$ and for a batch of $M$ opportunities, the constraint takes the form:
\begin{eqnarray} \label{eq:ra:batchcons}
\Expect\left[u_\alpha\left(\frac{1}{M} \sum_{i=1}^M W_i s(b(X_i),W_i)\right)\right] \geq u_\alpha(B) .
\end{eqnarray}
One can simplify this constraint based on the fact that the winning price $W_i$ and features $X_i$ are i.i.d. variables, and that $W$ is normally distributed given $X$. We refer the reader to Appendix \ref{app:rac:reduction} for detailed proof of the following lemma.

\begin{lemma}\label{thm::rac:reduction}
Constraint \eqref{eq:ra:batchcons} is equivalent to $\Expect[h(b(X),X)]\geq -1$, 
where
\begin{align}
    &h(b,X):=\notag\\
    &- e^{\gamma_1(X)} \Phi \left(\frac{b-\hat{w}(X)-\alphap\sigma(X)^2}{\sigma(X)}\right)-e^{\gamma_2}+e^{\gamma_2}\Phi \left(\frac{b-\hat{w}(X)}{\sigma(X)}\right)\label{eq:def_h}
\end{align}
with $\alphap:=\alpha/M$, $\gamma_1(X):=(1/2)(\alphap)^2\sigma(X)^2+\alphap\hat{w}(X)-\alphap B$, and $\gamma_2:=-\alphap B$. 
\end{lemma}

\begin{remark}
While the RMP approach \cite{Zhang_2017}  controls the risk of low profit by trading off between the mean and standard deviation of profit, which randomness is caused by CTR uncertainty, our method controls the risk of going over budget by measuring the expected utility of expenses, which randomness is caused by both the uncertainty of winning price distribution and the probability of winning the auction.
\end{remark}

\subsection{Optimal Solution for RAP Model}

Following the reductions presented in Sections \ref{sec:rabc}, we can reduce the problem to the following risk-averse expected instantaneous profit maximization problem:
\begin{eqnarray} \label{eq:rap}
\bRAP(\cdot) := \argmax_{b:\mathcal{X}\rightarrow \Re^+} &&\Expect[VCs(b(X),W)] - \Expect[Ws(b(X),W)] \nonumber\\
\subto&& \Expect[h(b(X),X)] \geq -1 .
\end{eqnarray}

In an attempt to solve this problem, one can again introduce the Lagrangian coefficient $\lambda\geq 0$ to obtain a relaxation of this risk-averse profit maximizing problem:
\begin{align*} 
\tbRAP_{\lambda}(\cdot) := \argmax_{b:\mathcal{X}\rightarrow \Re^+} \quad & \Expect[VCs(b(X),W)] - \Expect[Ws(b(X),W)] \\
& - \lambda (-1 - \Expect[h(b(X),X)]) \\
=\argmax_{b:\mathcal{X}\rightarrow \Re^+} \quad & \Expect[\Gain_{\lambda}(b(X),X)] ,
\end{align*}
where
\begin{align}
\Gain_{\lambda}(b,X) &:= 
 \hat{V}(X)\theta(X)\Phi\left(\frac{b-\hat{w}(X)}{\sigma(X)}\right) - g(b,X) 
+ \lambda(1 - h(b,X)), 
\end{align}
following our definitions of $g(b,X)$ and $h(b,X)$ in \eqref{eq:reduced:expense} and \eqref{eq:def_h} respectively.

The optimal bid price $\bRAP$ can be approximated using $\tbRAP_{\lambda^*}$
with $\lambda^*$ as the smallest value of $\lambda\geq 0$ such that 
$\Expect[h(\tbRAP_{\lambda^*}(X),X)] \geq -1$, where $\lambda^*$ can again be found using the bisection method.

We next provide a closed-form solution for $\tbRAP_\lambda$ in the form of Lemma \ref{thm:closedform:rap} (see complete proof in Appendix \ref{app:closedform:rap:proof}).

\begin{lemma}\label{thm:closedform:rap}
For any $\lambda\geq 0$, a maximizer of the Lagrangian relaxation takes the form:
\begin{align*}
& \forall X\in\mathcal{X} ,\;\tbRAP_{\lambda}(X):= \\
 &\quad \arg\max_{b\in \left\{0, \; - \frac{\mathbf{W}(\lambda \alphap e^{(\hat{V}(X)\theta(X)+\lambda e^{\gamma_2}-B) \alphap})}{\alphap }  + \hat{V}(X)\theta(X)+\lambda e^{\gamma_2}, \; \infty\right\}} \Gain_\lambda(b, X),
\end{align*}
where $\mathbf{W}$ is the Lambert W-function \cite{lambertw}, 
i.e. the inverse of $f(x):=x e^x$.
\end{lemma}
We note that given its analytical form, Lemma \ref{thm:closedform:rap} explicitly characterizes the influence of the customer value, budget, estimated CTR, and risk aversion level on the bid price, thus making the prescribed bid price highly interpretable.

\section{Experimental Set-up}\label{sec:exp}
To test the effectiveness of our bidding policies, we design our experiments using the real-life iPinyou dataset\footnote{\url{https://contest.ipinyou.com/}}. This dataset includes logs of ad biddings, impressions, clicks, and final conversions, while data are collected from different industries. Researchers in \cite{datasetbenchmark} have analyzed the distributions of data from different industries in this dataset and observed high variations between industries compared to within the same industry. Since the bidding models that we designed are based on the assumptions that each opportunity is i.i.d., we focus on one industry data: the Chinese vertical e-commerce industry collected from \textit{Advertiser ID} 1458.

The estimators $\theta(X), \hat{w}(X), \sigma(X)$ are trained and Lagrangian parameter $\lambda$ optimized using the \textbf{Training} set (3,083,056 observations), while other hyperparameters are tuned using the \textbf{Validation} set (307,319 observations). The \textbf{Test} set (307,319 observations) is used for out-of-sample (OOS) performance evaluation. During validation and test, the simulated batches ($M=10,000$ opportunities) are interrupted whenever the budget is fully utilized to mimic realistic practices. For OOS experiments, the risk aversion parameter $\alpha$ was fixed to the value that achieved, during validation, the highest Sharpe ratio of batch profit, i.e. 
\[\text{Sharpe ratio of profit} = \frac{\Expect[\text{Batch profit}]}{\sigma(\text{Batch profit})}\]
while preserving the \quoteIt{early stop frequency} lower than 5\%:
\[\text{Early stop frequency} = \mathbb{P} \left(\frac{\text{Batch expense}}{M} \geq B \right). \]

Finally, we investigate the role of the budget parameter $B$ on performance by considering $B\in\{\barB, \barB/2, \barB/4, \barB/8, \barB/16, \barB/32, \barB/64\}$ with $\barB$ as the historical average winning price in the dataset. This is in line with the ranges investigated in \cite{orbt,Zhang_2017}.

\section{Numerical Results}\label{sec:num}

In this section, we present our numerical results. We first investigate the effect of the risk aversion parameter on the capacity of the bidding policy to stay within budget. We then more formally compare the OOS performance of our proposed models to the model proposed in \cite{Zhang_2017}.

\subsection{RAP's Control of Budget Risk}
\begin{figure}
    \centering
    \includegraphics[width=0.8\textwidth]{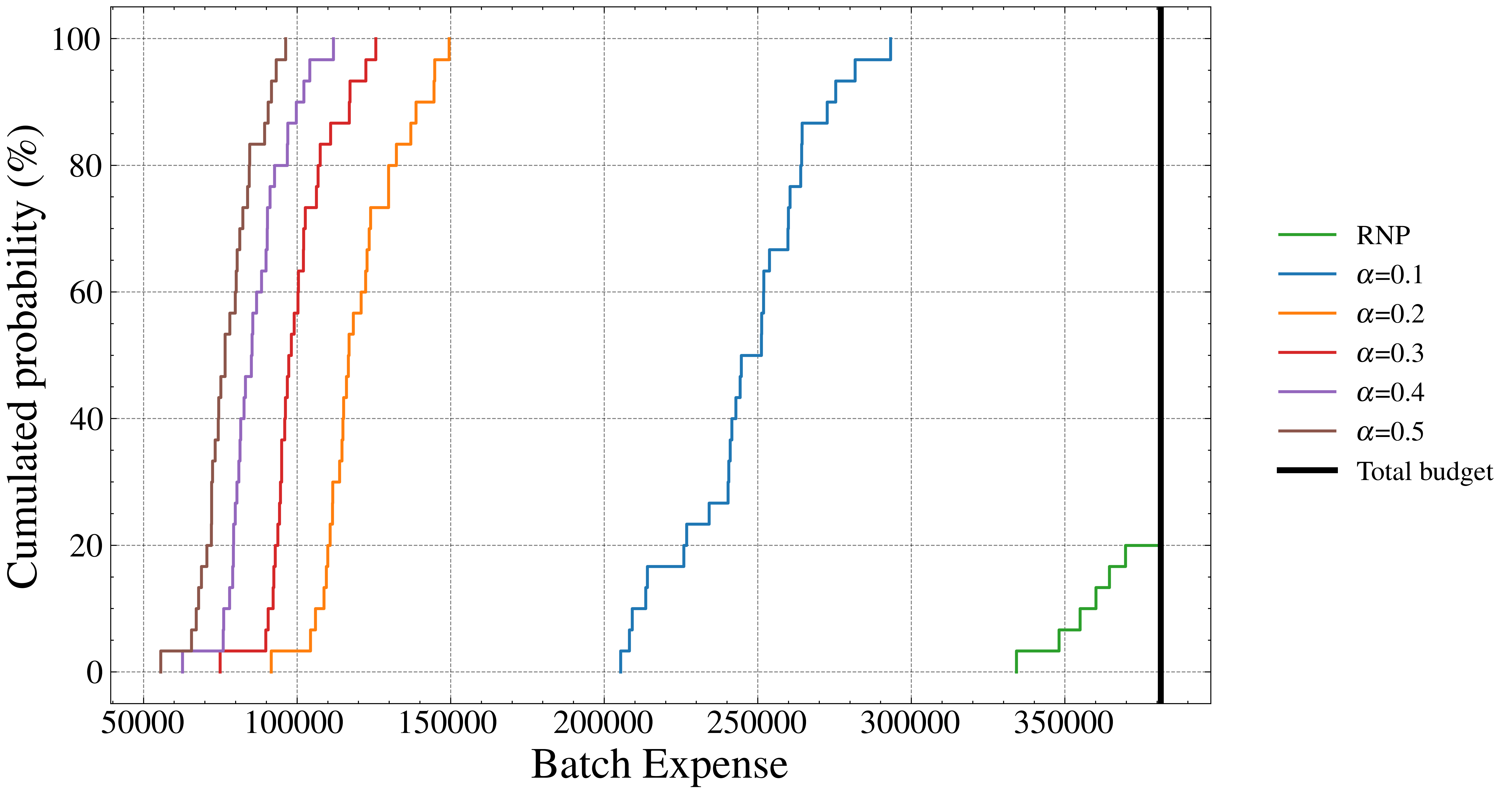}
    \caption{Empirical distribution (in validation) of batch expense for RAP, under different risk aversion levels, and RNP when B=1/2 $\barB$}
    \label{fig:profit_batchexpense}
\end{figure}

First, we investigated RAP's ability to control the budget risk in the validation set, through manipulation of the risk aversion parameter $\alpha$, and compared the performance to RNP. Figure \ref{fig:profit_batchexpense} presents the empirical cumulative distribution functions (CDF) of the batch expense obtained, for the different problem formulations when $B=\barB/2$. 
Compared with the RNP model, the risk-averse approach RAP demonstrates effective control of budget risk. Indeed, one can remark that all of the RAP models remain within total budget (black vertical line) for 100\% of the runs. This is not the case for RNP, which exceeds the budget 80\% of the time.

\begin{figure}
    \centering
    \includegraphics[width=0.8\textwidth]{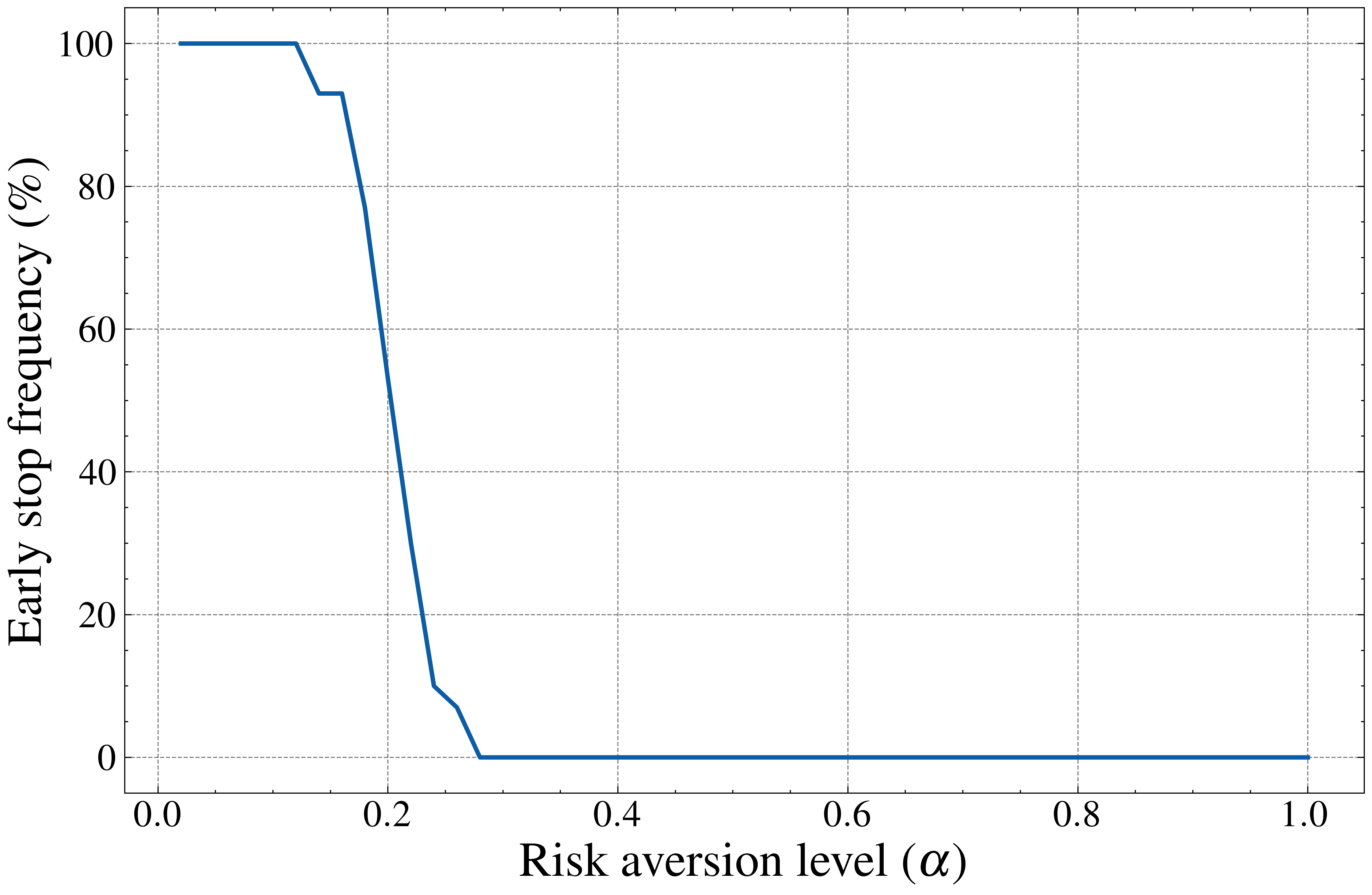}
    \caption{Empirical early stop frequency (in validation) under different risk levels for the profit model with $B=1/32 \barB$}
    \label{fig:alpha_stopfq_small}
\end{figure}

When reducing the marginal budget to $B=\barB/32$, the risk aversion parameter starts playing a more important role on budget risk. Indeed, Figure \ref{fig:alpha_stopfq_small} presents the empirical early stop frequency as a function of $\alpha$, where $\alpha$ needs to be greater than 0.28 for the bidding policy to have the guarantee to stay within budget. This confirms that RAP successfully handles the budget risk through the parametrization of $\alpha$.
%
%
%

We close this subsection with Table \ref{tab:Metrics}, which presents OOS performance metrics, on test data, for RNP and RAP, with risk aversion level selected based on validation data using Sharpe ratio and early stop frequency, under different levels of budget.
\begin{table*}
    \caption{RAP and RNP performance in test set} 
    \small
    \label{tab:Metrics}
    \centering
    \resizebox{\textwidth}{!}{
    \begin{tabular}{ |c|cc|cc|cc|cc|cc|cc| } 
    \hline
    \multirow{2}{*}{\textbf{Metrics}} &
      \multicolumn{2}{c|}{$\barB/2$} &
      \multicolumn{2}{c|}{$\barB/4$} &
      \multicolumn{2}{c|}{$\barB/8$} &
      \multicolumn{2}{c|}{$\barB/16$} &
      \multicolumn{2}{c|}{$\barB/32$} &
      \multicolumn{2}{c|}{$\barB/64$}\\
      & RAP & RNP & RAP & RNP & RAP & RNP & RAP & RNP & RAP & RNP & RAP & RNP  \\
    \hline
    Avg. batch clicks       
                & 5.600  & 6.367  
                & 5.067  & 3.300 
                & 2.633  & 1.700 
                & 0.967 & 1.433 
                & 0.333 & 1.267 
                & 0.333 & 1.133\\ 
    Avg. batch profit       
                & 192574  & 169873  
                & 260767  & 95035 
                & 156302  & 51845 
                & 49833   & 76413 
                & 20844   & 85811 
                & 22367   & 86183 \\ 
    Avg. batch expense       
                & 292121  & 381178  
                & 177766  & 190589 
                & 71620  & 95294 
                & 33835  & 47645 
                & 8006  & 23822 
                & 6484  & 11910\\ 
Avg. impression rate      
                &64.5\% & 69.6\%
                &47.1\% & 34.8\%
                &28.9\% & 18.4\%
                &19.4\% & 11.1\% 
                &7.7\% & 6.9\% 
                &6.8\% & 4.4\% \\
    Sharpe ratio of profit      
                & 1.083  & 0.847
                &1.381   & 0.595 
                &1.077   & 0.430 
                &0.497  & 0.702  
                & 0.450 & 0.802 
                & 0.480 & 0.732 \\ 

    Early stop frequency         
                & 0\%      & 100\% 
                &13.3\% & 100\%  
                & 0\%      & 100\% 
                & 0\%      & 100\%
                & 0\%      & 100\%  
                & 3.3\%      & 100\%
\\  [0.5ex] 
    \hline
    \end{tabular}}
\end{table*}
Based on this table, we first conclude by looking at the early stop frequency that the RAP controls better the risk of violating the budget constraint, at all budget levels, compared with the risk-neutral model. When the budget is relatively large, i.e. $B \geq \barB/8$, we also observe that the RAP outperforms the risk-neutral model RNP in Sharpe ratio of profit. When the budget is relatively small $B \leq \barB/16$, the RNP model generally outperforms the RAP models in terms of Sharpe ratio of profit and average batch clicks, whereas the RAP model has a better batch impression rate. This is because the bid price tends to be lower (i.e. more conservative) under the RAP. RAP is, therefore, able to invest in more ads thus getting more impressions. On the other hand, it is less competitive for the costly opportunities that end up generating more profit. 

\subsection{Comparison to RMP from \cite{Zhang_2017}}

\begin{figure*}
\centering

\subfloat[$B=\barB/2$]{
  \includegraphics[width=0.42\textwidth]{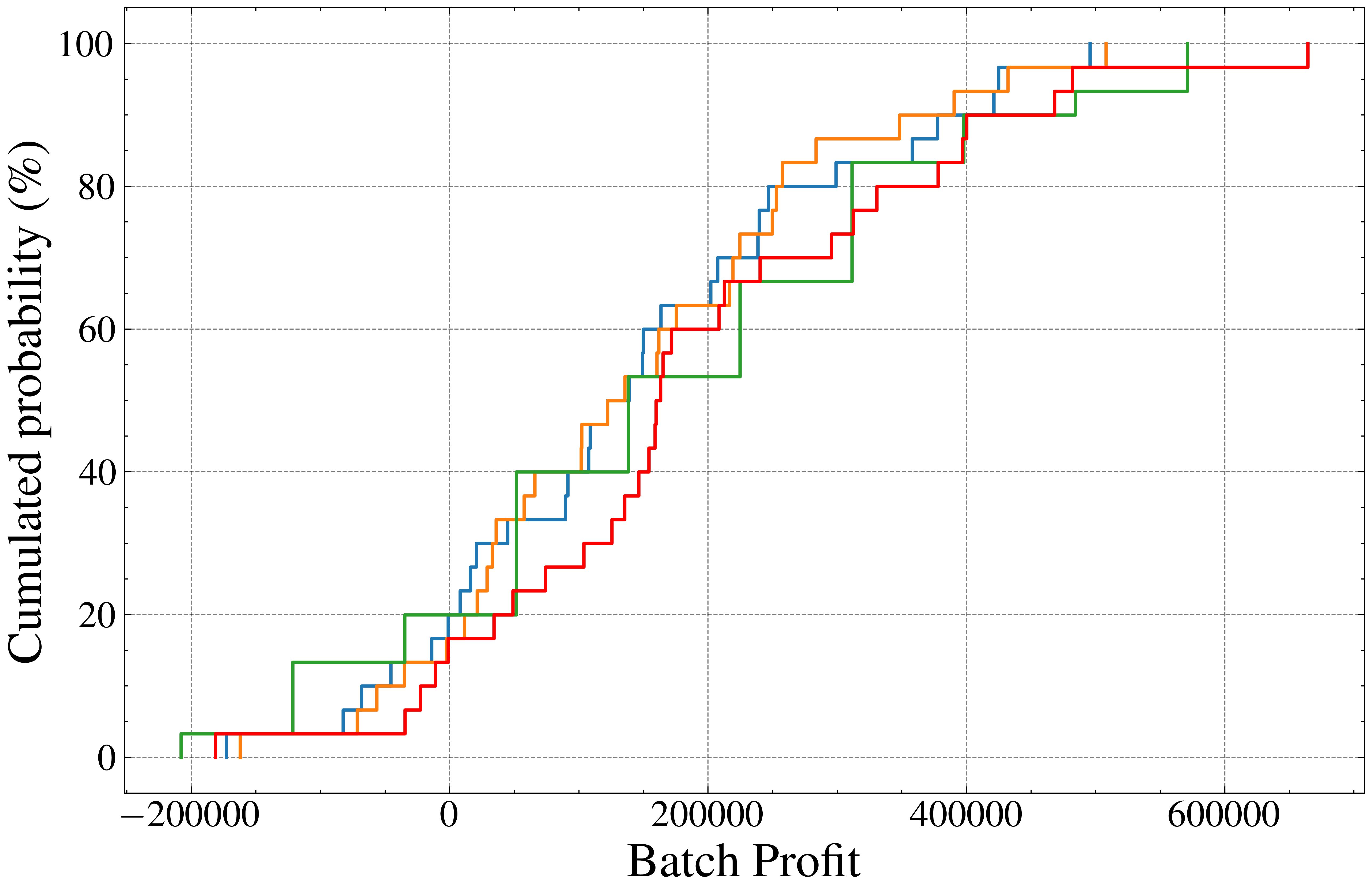}
}
\subfloat[$B=\barB/32$]{
  \includegraphics[width=0.5\textwidth]{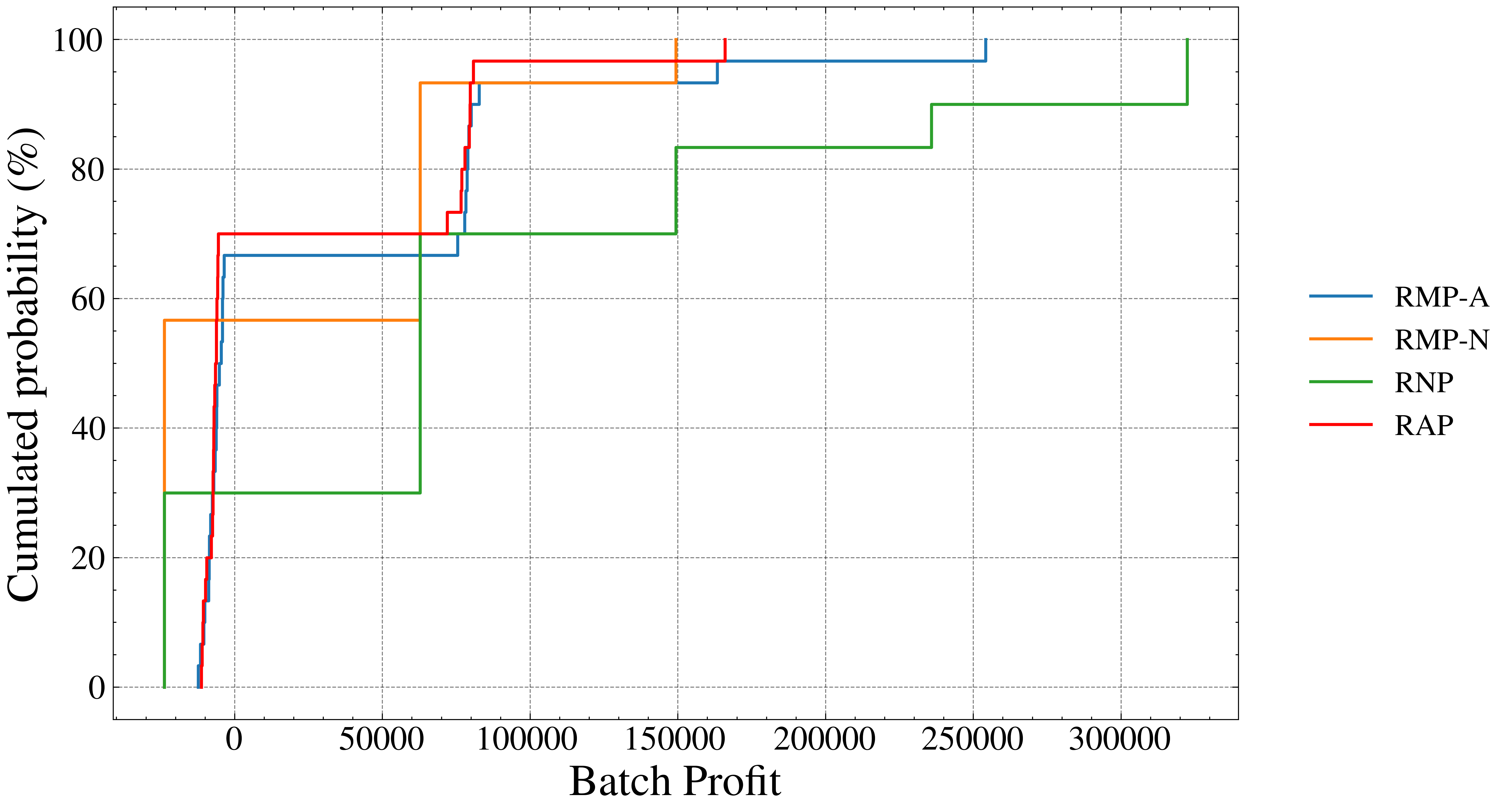}
}
\caption{Out-of-sample empirical distribution of batch profit under two different budget levels.}\label{fig:compare_batchprofit}
\end{figure*}

\begin{figure*}
\centering

\subfloat[$B=\barB/2$]{
  \includegraphics[width=0.415\textwidth]{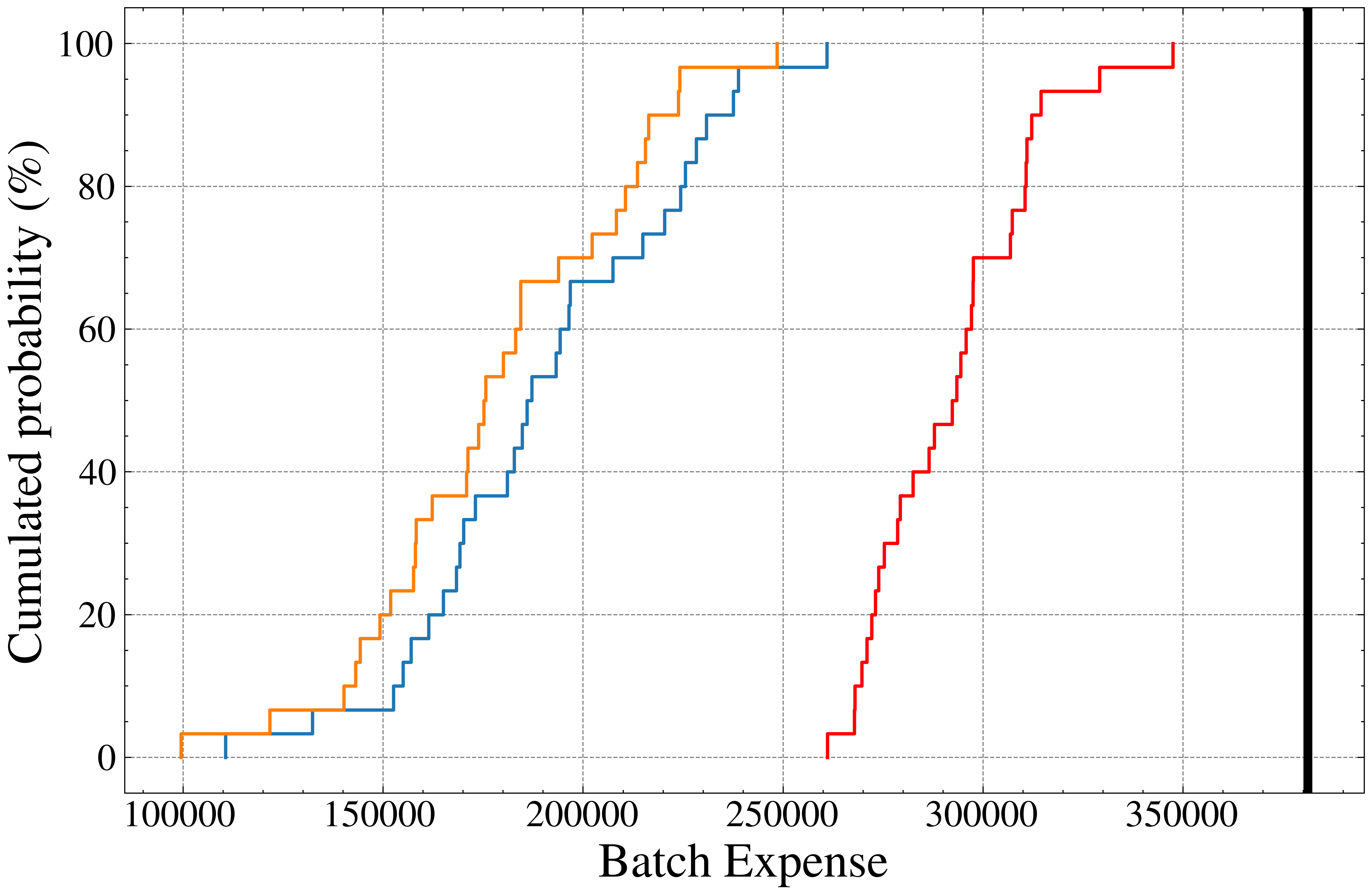}
}
 \vspace{0.5mm}
\subfloat[$B=\barB/32$]{
  \includegraphics[width=0.52\textwidth]{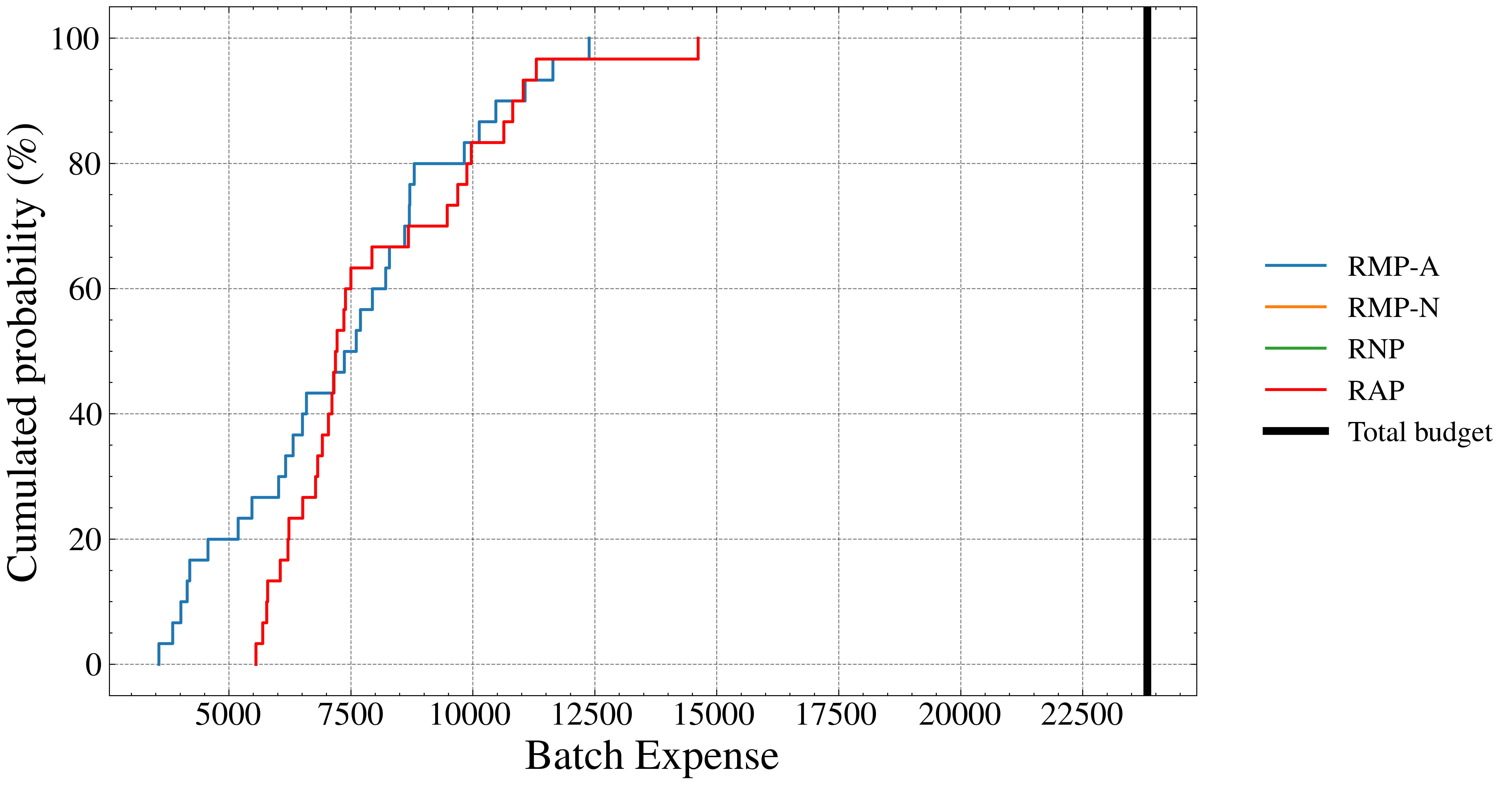}
}
\caption{Out-of-sample empirical distribution of batch expense under two different budget levels. Note that curves are not presented when they overlap with the total budget bar.}\label{fig:compare_batchexpense}
\end{figure*}

We now compare the performance of RAP and RNP on the test set to the risk-neutral (called RMP-N) and risk-averse (called RMP-A) bidding policy obtained from the RMP proposed in \cite{Zhang_2017}. To offer a fair comparison, we select hyperparameters in RMP using the same procedure as for RAP. Table \ref{tab:compare_rmp} presents the different metrics for the OOS performance of the different approaches under two different levels of budget (i.e. high and low). 

Regarding the control of budget risk, we can observe, based on a 0\% early stop frequencies, that both the RAP and RMP-A approaches produce bid policies that control well this risk. This is not the case for RNP and RMP-N, which are by design risk-neutral. In fact, RMP-N appears to be slightly over-conservative at higher budget levels with an early stop frequency of 0\%. It is confirmed in Figure \ref{fig:compare_batchexpense}, which presents the CDFs of batch expense, and where RMP-N's curve lands a significant margin away from the budget mark for $B=\barB/2$. 

Next, we observe that RNP always outperforms both RMP-A and RMP-N in terms of average batch profit. This is especially noticeable at the low budget level where the average batch profit is 2.8 times larger than what is achieved by the best competing approach. Figure \ref{fig:compare_batchprofit} presents the CDFs of batch profit. In the case of the high budget (see Figure \ref{fig:compare_batchprofit}(a)), the CDFs appear to mostly overlap with a slight trend to the right for both RNP and RAP, which explains the better average performances. We note that the RNP also appears to have a heavier left tail, which would indicate that it exposes the decision maker to a slightly larger risk of a net loss. In contrast, RAP appears protected from these losses and achieves a larger average batch profit than RNP. We explain the better performance of RAP over RNP here by the fact that when the budget is large, the optimal risk aversion parameter can be reduced in favor of improved expected profit. Moreover, it appears that a small level of risk aversion might actually play the role of making the policies performance more robust out-of-sample, i.e. reducing generalization error. Note that the observations made about average profit can also be made about the Sharpe ratio, identifying our RAP as a clear winner for a higher budget and our RNP as a clear winner for a lower budget.

Finally, in terms of impression rates, we observe that while all methods appear to perform similarly at the low budget level, both RAP and RNP clearly outperform RMP-A/N models at the higher budget level. We interpret this as evidence that our approach is more successful at predicting the price at which an opportunity will be won due to our modeling of the conditional influence of $X$ on the distribution of $W$.

\begin{table*}
    \caption{Performance comparison between our RAP and RNP, and the RMP-N and RMP-A from \cite{Zhang_2017} } 
    \label{tab:compare_rmp}
    \centering
    \resizebox{\textwidth}{!}{\begin{tabular}{ |c||cc|cc||cc|cc| } 
    \hline
    \multirow{2}{*}{\textbf{Metrics}} &
      \multicolumn{4}{c||}{$\barB/2$} &
      \multicolumn{4}{c|}{$\barB/32$} \\
      \cline{2-9}
      & RAP & RNP & RMP-A & RMP-N & RAP & RNP & RMP-A & RMP-N  \\
    \hline
    Avg. batch clicks       & 5.600  & \textbf{6.367}   & 3.867  & 3.700  
                     & 0.333 & \textbf{1.267}   & 0.433 & 0.500  \\ 
    Avg. batch profit       & \textbf{192574}  & 169873   & 144471  & 142193 
                     & 20844   & \textbf{85811}   & 30136 & 19454  \\ 
    Avg. batch expense       & 292121  & 381178   & 190200  & \textbf{178052}  
                     & 8006  & 23822   & \textbf{7370} & 23822  \\ 
    Avg. impression rate      &64.5\% & \textbf{69.6\%} &41.5\%  & 38.8\% 
                             &\textbf{7.7\%} & 6.9\% &5.0\% & 7.3\% \\ 
    Sharpe ratio of profit    & \textbf{1.083}  & 0.847  & 0.892 & 0.922 
                               & 0.450 & \textbf{0.802}  & 0.487 & 0.363 \\ 
    Early stop frequency      & 0      & 100\%   & 0      & 0        
                              & 0      & 100\%   & 0      & 100\% \\  [0.5ex] 
    \hline
    \end{tabular}}
\end{table*}

\section{Conclusion}\label{sec:conc}
In this paper, we proposed an approach for optimizing bid policies in a context where one wishes to control the risk of spending more than a given budget in a certain period of time (captured by a total number of opportunities $M$). We proposed both a risk-averse and a risk-neutral problem that let the decision maker control how much protection is needed against overspending. We further derived closed-form expressions for the optimal bid policies when using a mixed stochastic model, which employed the dataset's empirical distribution for the type of opportunities $X$ and trained CTR and conditional Gaussian distribution models for the clicks and winning prices respectively. The closed-form expressions benefit from being interpretable and easy to implement in production. Our two approaches were then compared to analogous risk-neutral and risk-averse bid policies from the RMP approach in \cite{Zhang_2017} using the iPinyou dataset. The results provide empirical evidence that RAP and RNP provide significant improvement in generating high profit while controlling the risk of going over budget. 

In terms of future work, one could consider accounting for risk aversion with respect to profit generation, i.e. $\Expect[u(\text{Batch profit})]$. Alternatively, one could investigate how the choice of alternate convex risk measures, such as conditional value-at-risk, might affect the closed-form solutions and performance. Finally, many other aspects of bid optimization could be handled using similar risk measures: e.g. risk of not reaching a batch revenue, impression rate, or total click target. Finally, one could explore other types of implementations for our bidding policy. A shrinking horizon scheme could be used where the bidding policy continuously accounts for the budget that is left to invest in new ads. Alternatively, a policy optimization approach could be used to optimize the parameters of our policy in an RL environment. 

\section*{Acknowledgement}
This research was supported financially by the NSERC (RGPIN-2016-05208), the Fin-ML CREATE program, MITACS' RTA program, and 
enabled in part by support provided by the Digital Research Alliance of Canada.

\appendix

\section{Proofs}
\subsection{Proof of Lemma \ref{thm:closedform:rnp}\label{app:rnp:thm:proof}}

We start by exploiting the interchangeability property of expected value (see \cite{Shapiro:interchange}), which implies that the optimal bid price for the Lagrangian relaxation can be obtained as the price, for each $X$, that maximizes the Lagrangian relaxation function $\Gain_\lambda(b,X)$ 
with
\begin{align*}
\Gain_\lambda&(b,X)\\&:=\hat{V}\theta\Phi\left(\frac{b-\hat{w}}{\sigma }\right) -(\lambda+1)\left[\hat{w}\Phi\left(\frac{b-\hat{w}}{\sigma}\right)- \sigma \phi\left(\frac{b-\hat{w}}{\sigma}\right)\right] + \lambda B \nonumber \\
&= - (\hat{w} + \lambda \hat{w}-\hat{V}\theta)\Phi \left(\frac{b-\hat{w}}{\sigma}\right) + (1+\lambda) \sigma\phi\left(\frac{b-\hat{w}}{\sigma}\right) + \lambda B\,,
\end{align*}
where we dropped the relation to $X$ for simplicity of presentation. 

Since $\Gain_{\lambda}(b,X)$ is twice differentiable with respect to $b$, the maximizer for $b$ is either $0, \infty$ or at a value where the derivative is $0$. For the latter case, we get that:
\begin{align*}
&\frac{d \Gain_{\lambda}(b,X)}{db}=0  \Leftrightarrow\\
&\frac{\hat{w}+\lambda \hat{w}-v\theta}{\sigma}\phi \left(\frac{b-\hat{w}}{\sigma}\right)+(1+\lambda) \frac{b-\hat{w}}{\sigma} \phi\left(\frac{b-\hat{w}}{\sigma} \right)=0\,.
\end{align*}
Hence, we can conclude that the value of $b$ where $\Gain_{\lambda}(b, X)$ has a derivative of zero is $b=\hat{V}(X)\theta(X)/(\lambda+1)$.\qed

\subsection{Proof of Lemma \ref{thm::rac:reduction}\label{app:rac:reduction}}

We can simplify constraint \eqref{eq:ra:batchcons} based on the fact that the winning price $W_i$ and features $X_i$ are i.i.d. variables:
\begin{align*}
&\Expect \left[u_\alpha\left(\frac{1}{M} \sum_{i=1}^M W_i s(b(X_i),W_i)\right) \right] = -\Expect\left[ e^{\alpha \left( \frac{1}{M} \sum W_i s(b(X_i),W_i)\right)}\right] \\
&\quad= - \Expect\left[\prod_{i=1}^M e^{(\alpha/M) W_i s(b(X_i),W_i)}\right] = -\prod_{i=1}^M \Expect\left[ e^{(\alpha/M) W_i s(b(X_i),W_i)}\right] \\
&= -\Expect\left[ e^{\alpha/M W s(b(X),W)}\right]^M= \Expect\left[u_{\alpha/M}( W s(b(X),W)\right]^M,
\end{align*}
where the second equality is derived based on the independence assumption and the common distribution.

It means that constraint \eqref{eq:ra:batchcons} can be rewritten as: 
%
\begin{eqnarray} \label{eq:ra:cons}
\Expect[u_{\alphap} ( W s(b(X),W))] \geq u_{\alphap}(B) ,
\end{eqnarray}
where $\alphap := \frac{\alpha}{M} $.

We further simplify the constraint to obtain a closed-form representation. In doing so, we start by dividing both side of constraint \eqref{eq:ra:cons} by $-u_{\alphap}(B)>0$ in order to normalize this constraint. Note that $B \geq 0$ implies that $-u_{\alphap}(B)>0$. This leads us to
\[-\Expect[u_{\alphap}(Ws(b(X),W))/u_{\alphap}(B)] \geq -1 .\]

We then exploit the Gaussian nature of $W$ when $X$ is known to obtain the following reduction:
\begin{align} \label{eq:rap:utility_reduce}
& -\Expect[u_{\alphap}(W s(b(X),W))/u_{\alphap}(B)|X] \nonumber \\
&= -e^{-\alphap B}\left(\int_{w\leq b(X)} \exp(\alphap w)f_{W|X}(w) dw \right.\notag\\
&\quad\quad\quad\quad\quad\quad\quad\left.+ \int_{w> b(X)} \exp(\alphap \cdot 0) f_{W|X}(w) dw \right) \nonumber\\
&= -e^{-\alphap B}\left(\int_{w\leq b} \frac{1}{\sigma\sqrt{2\pi}}\exp\left(- \frac{(w-\hat{w})^2}{2\sigma^2} + \alphap w\right) dw +1-\Phi\left(\frac{b-\hat{w}}{\sigma}\right)\right) \nonumber\\
&= -e^{-\alphap B}\left(\int_{w\leq b} \frac{1}{\sigma\sqrt{2\pi}}\exp\left(- \frac{(w - (\hat{w} + \alphap \sigma^2))^2}{2\sigma^2}\right.\right.\notag\\
&\quad\quad\quad\quad\quad\quad\quad\left.\left.+\frac{(\alphap)^2 \sigma^2}{2} + \alphap \hat{w}\right) dw +1 -\Phi\left(\frac{b-\hat{w}}{\sigma}\right)\right) \nonumber \\
&= -e^{\gamma_1} \int_{w\leq b} \frac{1}{\sigma\sqrt{2\pi}}\exp\left(- \frac{(w-\hat{w}-\alphap\sigma^2)^2}{2\sigma^2}\right) dw-e^{\gamma_2}+e^{\gamma_2}\Phi\left(\frac{b-\hat{w}}{\sigma}\right) \nonumber \\
&=  - e^{\gamma_1(X)} \Phi \left(\frac{b(X)-\hat{w}(X)-\alphap\sigma(X)^2}{\sigma(X)}\right)-e^{\gamma_2}+e^{\gamma_2}\Phi \left(\frac{b(X)-\hat{w}(X)}{\sigma(X)}\right)\notag
\end{align}

\removed{\begin{align} \label{eq:rap:utility_reduce}
& e^{\gamma_2(X)} \Expect[u_{\alphap}(W s(b(X),W))/u_{\alphap}(B)|X] \nonumber \\
&= \int_{w\leq b(X)} \exp(\alphap w)f_{W|X}(w) dw \notag\\
&\quad\quad\quad\quad\quad\quad\quad+ \int_{w> b(X)} \exp(\alphap \cdot 0) f_{W|X}(w) dw \nonumber\\
&= \int_{w\leq b} \frac{1}{\sigma\sqrt{2\pi}}\exp\left(- \frac{(w-\hat{w})^2}{2\sigma^2} + \alphap w\right) dw +1-\Phi\left(\frac{b-\hat{w}}{\sigma}\right) \nonumber\\
&= \int_{w\leq b} \frac{1}{\sigma\sqrt{2\pi}}\exp\left(- \frac{[w - (\hat{w} + \alphap \sigma^2)]^2}{2\sigma^2}\right.\notag\\
&\quad\quad\quad\quad\quad\quad\quad\left.+\frac{(\alphap)^2 \sigma^2}{2} + \alphap \hat{w}\right) dw +1 -\Phi\left(\frac{b-\hat{w}}{\sigma}\right) \nonumber \\
&= \int_{w\leq b} \frac{1}{\sigma\sqrt{2\pi}}\exp\left(- \frac{(w-\hat{w}-\alphap\sigma^2)^2}{2\sigma^2}\right) dw-e^{\gamma_2}+e^{\gamma_2}\Phi\left(\frac{b-\hat{w}}{\sigma}\right) \nonumber \\
&=   \Phi \left(\frac{b(X)-\hat{w}(X)-\alphap\sigma(X)^2}{\sigma(X)}\right)-e^{\gamma_2}+e^{\gamma_2}\Phi \left(\frac{b(X)-\hat{w}(X)}{\sigma(X)}\right)\notag
\end{align}}
where we temporarily drop the relation to $X$ for simplicity. 
This completes our proof. \qed

\subsection{Proof of Lemma \ref{thm:closedform:rap}\label{app:closedform:rap:proof}}
Similar to the risk-neutral models, we also exploit the interchangeability property of expected value, which implies that the optimal bid price for the Lagrangian relaxation can be obtained as the price, for each $X$, that maximizes the Lagrangian relaxation function $\Gain_\lambda(b, X)$:  
\begin{align*}
\Gain_{\lambda}(b,X) &:= (\hat{V}\theta+\lambda e^{\gamma_2})\Phi\left(\frac{b-\hat{w}}{\sigma}\right) - \lambda e^{\gamma_1} \Phi\left(\frac{b-\hat{w}-\alphap\sigma^2}{\sigma}\right)  \nonumber\\
& \quad\quad\quad\quad+ (1-e^{\gamma_2})\lambda - \left[\hat{w}\Phi\left(\frac{b-\hat{w}}{\sigma}\right)- \sigma \phi\left(\frac{b-\hat{w}}{\sigma}\right)\right] \nonumber \\
 &= c_1\Phi\left(\frac{b-\hat{w}}{\sigma}\right) - c_2 \Phi\left(\frac{b-c_3}{\sigma}\right) + (1-e^{\gamma_2})\lambda + \sigma \phi\left(\frac{b-\hat{w}}{\sigma}\right) \,,
\end{align*}
where we again drop the dependence on $X$ for convenience and where we use $c_1$, $c_2$, and $c_3$ to refer to $c_1(X):=\hat{V}(X)\theta(X)+\lambda e^{\gamma_2} - \hat{w}(X)$, $c_2(X):=\lambda e^{\gamma_1(X)}$, and $c_3(X):=\hat{w}(X)+\alphap\sigma(X)^2$.

Since $\Gain_{\lambda}(b,X)$ is differentiable with respect to $b$, the maximizer for $b$ is either $0, \infty$ or at a value where the derivative is $0$. For the latter case, we get that:
\begin{align*}
& \quad \frac{\mathrm{d} \Gain_{\lambda}(b,X)}{\mathrm{d} b} =0 \\
&\Leftrightarrow - \frac{c_1}{\sigma}\phi\left(\frac{b-\hat{w}}{\sigma}\right) +  \frac{c_2}{\sigma} \phi\left(\frac{b-c_3}{\sigma}\right) + \frac{b-\hat{w}}{\sigma}\phi\left(\frac{b-\hat{w}}{\sigma}\right)= 0\\
&\Leftrightarrow (c_1-b+\hat{w})\phi\left(\frac{b-\hat{w}}{\sigma}\right) = c_2 \phi\left(\frac{b-c_3}{\sigma}\right) \\
&\Leftrightarrow \ln\left((\hat{V}\theta+\lambda e^{\gamma_2}-b)\phi\left(\frac{b-\hat{w}}{\sigma}\right)\right) = \ln \left(c_2 \phi\left(\frac{b-c_3}{\sigma}\right)\right) \\
&\Leftrightarrow \ln\left(\frac{\hat{V}\theta+\lambda e^{\gamma_2}-b}{\sqrt{2\pi}}\right)-\frac{(b-\hat{w})^2}{2\sigma^2} = \ln\left(\frac{c_2}{\sqrt{2\pi}}\right)-\frac{(b-c_3)^2}{2\sigma^2}\\
&\Leftrightarrow 2\sigma^2\ln(\hat{V}\theta+\lambda e^{\gamma_2}-b)-(b-\hat{w})^2 =\notag\\
&\quad\quad\quad\quad\quad\quad\quad\quad2\sigma^2(\ln(\lambda)+\gamma_1)-c_3^2 +2bc_3-b^2\\
&\Leftrightarrow 2\sigma^2\ln(\hat{V}\theta+\lambda e^{\gamma_2}-b)-(b-\hat{w})^2 =\notag\\ &\quad\quad\quad\quad\quad\quad2\sigma^2[\ln(\lambda)+\gamma_1]-(\hat{w}+\alphap \sigma^2)^2 + 2b(\hat{w}+\alphap \sigma)-b^2\\
&\Leftrightarrow 2\ln(\hat{V}\theta+\lambda e^{\gamma_2}-b) = 2(\ln(\lambda)+0.5(\alphap)^2\sigma^2+\alphap\hat{w}-\alphap B)\notag\\
&\quad\quad-(\alphap)^2\sigma^2 - 2w\alphap + 2b\alphap \\
&\Leftrightarrow \ln\left(\frac{\hat{V}\theta+\lambda  e^{\gamma_2}-b}{\lambda}\right) = \alphap (b-B)\\
&\Leftrightarrow \ln\left(\frac{\hat{V}\theta+\lambda  e^{\gamma_2}-b}{\lambda}\right) = (\hat{V}\theta+\lambda   e^{\gamma_2}-B) \alphap - (\hat{V}\theta+\lambda  e^{\gamma_2}-b)\alphap  \\
&\Leftrightarrow \frac{\hat{V}\theta+\lambda  e^{\gamma_2}-b}{\lambda} e^{(\hat{V}\theta+\lambda  e^{\gamma_2}-b)\alphap}  = e^{(\hat{V}\theta+\lambda  e^{\gamma_2} -B) \alphap} \\
&\Leftrightarrow (\hat{V}\theta+\lambda  e^{\gamma_2}-b) e^{(\hat{V}\theta+\lambda  e^{\gamma_2}-b)\alphap}  = \lambda e^{(\hat{V}\theta+\lambda  e^{\gamma_2} - B) \alphap} \\
&\Leftrightarrow (\hat{V}\theta+\lambda  e^{\gamma_2}-b)\alphap e^{(\hat{V}\theta+\lambda  e^{\gamma_2}-b)\alphap}  = \lambda\alphap e^{(\hat{V}\theta+\lambda  e^{\gamma_2} - B) \alphap} \\
&\Leftrightarrow (\hat{V}\theta+\lambda e^{\gamma_2}-b)\alphap  = \mathbf{W}\left(\lambda \alphap e^{(\hat{V}\theta+\lambda e^{\gamma_2} - B) \alphap}\right) \\
&\Leftrightarrow b = - \frac{\mathbf{W}\left(\lambda \alphap e^{(\hat{V}\theta+\lambda e^{\gamma_2}-B) \alphap}\right)}{\alphap }  + \hat{V}\theta+\lambda e^{\gamma_2} \;,
\end{align*}
where the $\mathbf{W}(\cdot)$ is the Lambert-$\mathbf{W}$ function. This completes our proof. \qed

\bibliographystyle{apalike}
\bibliography{sample-base}

\end{document}